\documentclass[10pt]{article}
\usepackage{subcaption}
\usepackage{amsmath}
\usepackage{booktabs}
\usepackage{multirow}
\usepackage{siunitx}   
\usepackage{xcolor} 
\usepackage{tcolorbox}
\usepackage{stfloats}

\tcbuselibrary{skins,breakable}
\usepackage[letterpaper]{geometry}
\usepackage{hicss}
\usepackage{times}
\usepackage{url}
\usepackage{latexsym}
\usepackage{indentfirst}
\usepackage{graphicx}
\graphicspath{{images/}}
\usepackage[
    style=numeric,
  ]{biblatex}
\title{Evaluating Large Language Models for Forced Outage Risk Prediction: Benefits and Comparison to Machine Learning}

\author{Christos Petridis \\
 Temple University \\
 {\underline{christos.petridis@temple.edu}} \\ \And
 Zoran Obradovic \\
 Temple University \\
 {\underline{zoran.obradovic@temple.edu}} \\ \And
 Mladen Kezunovic \\
 Texas A\&M University \\
 {\underline{kezunov@ece.tamu.edu}} \\ }

\date{}

\begin{document}
\maketitle

{\renewcommand{\thefootnote}{}\footnotetext{This paper has been accepted at the 60th Hawaii International Conference on System Sciences (HICSS-60).}}

\begin{abstract}

This study examines the ability of large language models (LLMs) to predict the risk of weather-related forced outages in the distribution grid in a zero-shot framework, without labeled training data. The problem is formulated as a binary severity classification task across three forecast horizons (3h, 6h, 12h), using six years of outage records and high-resolution weather data for a utility service area in central Texas. Four zero-shot LLMs are benchmarked against two supervised classifiers across two input configurations: one using current weather observations and the other using weather forecast data. Results show that supervised models outperform LLMs on macro-F1 and precision, while newer LLM generations achieve competitive scores. Beyond accuracy, LLMs offer complementary strengths in actionable reasoning and geographic scalability, suggesting that combining them with supervised models may be the best practice.
\end{abstract}

\subsubsection*{Keywords:}

Forced Outage Prediction, Large Language Models, Zero-Shot Classification, Weather-Driven Outages, Distribution Grid Management

\section{Introduction}
\label{sec:intro}

Weather-related power outages are among the most enduring and significant threats to grid resilience. Approximately 83\% of reported major power outages in the United States between 2000 and 2021 are attributable to weather-related phenomena \parencite{doe2023keeping}, including thunderstorms, strong winds, flash floods, and winter storms. Annually, millions of customers are impacted, resulting in considerable economic costs for utilities and the communities they serve.

Prediction of weather-driven power outages using Machine Learning (ML) has received growing attention over the past decade. Early work focused on storm-level outage count regression, predicting the number of customers affected during a specific storm event by combining Numerical Weather Prediction (NWP) outputs with vegetation and soil data across a spatial grid of service locations \parencite{cerrai2019predicting}. While effective within their scope, these approaches are inherently event-triggered — they require a storm to be identified and characterized in advance, limiting their applicability to continuous, day-to-day operational decision-making. Subsequent studies scaled these approaches and explored Deep Learning (DL) ensembles for large-scale grid outage prediction \parencite{prieto2025predicting}, while others examined fault prediction under specific meteorological conditions such as heatwaves \parencite{atrigna2023machine}. Some prediction frameworks have proposed statistical models that use historical weather data to predict the number of power distribution interruptions in a region, with the goal of feeding the forecasts into smart-grid self-healing schemes to improve grid resilience  \parencite{sarwat2016weather}. More recently, a graph neural network approach with contrastive learning has been proposed to capture spatial dependencies among service locations during storm events \parencite{shen2025empowering}, though this method remains event-triggered, as it requires a storm event to be identified first.

A more operationally meaningful line of work frames outage prediction as a severity classification problem. A Conditioned Outage Prediction Model was proposed that classifies storm events into low-, moderate-, and high-severity groups and then trains a dedicated model on the severity-matched subset \parencite{yang2020enhancing}.
Recent work has approached outage risk prediction from several angles. A granular spatio-temporal grid-based DL framework was developed that discretizes the service territory into clusters and predicts forced outages jointly across space and time \parencite{daniel_hicss}. The sensitivity of such models to feature selection and algorithm choice was analyzed, revealing that predictive performance varies substantially across configurations \parencite{baembitov2024sensitivity}. An interpretable ML-based outage risk prediction framework argues that utilities will only adopt these models in practice if the predicted risk can be clearly tied to the factors causing it \parencite{rashid_interpretability}. These directions were consolidated into a unified state-of-risk formulation that jointly handles vegetation- and weather-related outages in distribution networks \parencite{baembitov2023state}. The aforementioned approaches frame the task as a binary problem of zero versus one or more outages. This method does not characterize the severity of outages, limiting operators' ability to deploy mitigation measures.

Motivated by prior work demonstrating that LLMs can predict structured numerical outcomes with competitive performance \parencite{gruver2023large}, we explore this approach in the context of forced outage risk prediction. We present what is believed to be the first application of LLM-based zero-shot inference to this problem. Our approach offers three contributions: a) multiple design benefits over ML approaches, evaluated along the criteria of transferability, scalability, and interpretability, b) a significant reduction in implementation cost by eliminating the need for labeled outage data, complex feature engineering, and model training; and c) a comparative performance evaluation against supervised ML models, providing empirical grounding for understanding how LLMs can complement traditional approaches in operational settings.


The rest of the paper is organized as follows: Section ~\ref{sec:motivation} provides the motivation for using LLMs for weather-related forced-outage risk estimation. Section ~\ref{sec:statistical_analysis} demonstrates a statistically significant correlation between severe weather events and outages at the distribution grid. Section ~\ref{sec:FramingOutagePredictionSeverity} frames the outage risk prediction as a severity classification problem. Section ~\ref{sec:data_model} describes the data modeling, and Section ~\ref{sec:predictive_model} presents the predictive modeling setup. Section ~\ref{sec:experimental_results} reports the experimental results obtained using historical real-world outage data. Section ~\ref{sec:discussion} discusses the findings, and Section ~\ref{sec:conclusion} provides the conclusions, followed by references and an appendix.



\section{Motivation for Large Language Models}
\label{sec:motivation}
Prior research has demonstrated that ML/DL models can predict weather-driven outages with considerable accuracy, particularly when combining multiple data sources, including weather forecasts, vegetation indices, and spatial grid features. However, these approaches incur high operational costs, requiring extensive data wrangling and careful feature engineering, large volumes of labeled historical outage data, and periodic retraining as grid conditions evolve (data distribution shift). As an illustration, the ML pipeline in this work required aligning feeder-level outage records with weather data from a 10×10 grid, addressing severe class imbalance ($\approx$5\% positive rate) through upweighting and regularization, and validating via Leave-One-Year-Out cross-validation across six years of data. Each of these steps requires specialized ML expertise, and when deployed to a new service territory, the entire process must be repeated from scratch, introducing limitations with transferability. This raises a practical question of whether other approaches could serve as viable alternatives in settings where such resources are unavailable or expensive to implement and maintain. For the various reasons discussed further below, we hypothesized that the LLM approach may offer significant benefits at a lower cost overall, and with acceptable performance.

To formally evaluate the benefits of LLM approaches, as a viable alternative, when compared to the traditional (ML) approaches, we further hypothesized that LLMs may offer advantages along the three important design criteria that directly affect the cost of the solution: a) Transferability, which evaluates the solution’s ability to be developed in one utility setting and then be applied to other utility settings, b)  Scalability, which evaluates the solution’s ability to be adjusted to any spatiotemporal operational and geographic conditions, and c) Interpretability, which evaluates how easy it is for the utility staff to interpret the severity of the outage impact. The justification for the hypothesis is the fact that: a) LLMs are based on their pre-trained knowledge of meteorology and infrastructure risk for any selected time frame or territory, b) LLMs provide a result that may be crafted for a particular time or space scalability through the formulation of the prompts, and c) LLM predictions, most importantly, can be accompanied by a natural-language explanation of the key factors driving risk assessment. For example, in our experiments, the model produced reasoning such as \textit{``Heavy rain, high precipitation, and strong wind gusts up to 52.9\,km/h significantly increase outage risk''} for a high-risk prediction, and \textit{``Wind gusts are moderately strong but precipitation and severe weather indicators are low, so widespread outages are unlikely''} for a low-risk one, though the level of detail can be adjusted to the needs of the deployment. Our prompt instructs the model to provide a short sentence of no more than 30 words explaining the main contributing factors (see Section \ref{sec:PredictiveLLM}). This kind of output is immediately actionable and requires no technical interpretation by the operator. A central question in this work is therefore whether, besides the benefits in transferability, scalability and interpretability, LLMs can accurately identify high-risk conditions from weather inputs alone (without prior knowledge of historical grid outages or of the spatiotemporal scale relevant to different utility staff), thereby enabling staff to make informed decisions in any utility setting.

\section{Statistical Analysis}
\label{sec:statistical_analysis}

\begin{figure}[b]
  \centering
  \includegraphics[width=\linewidth]{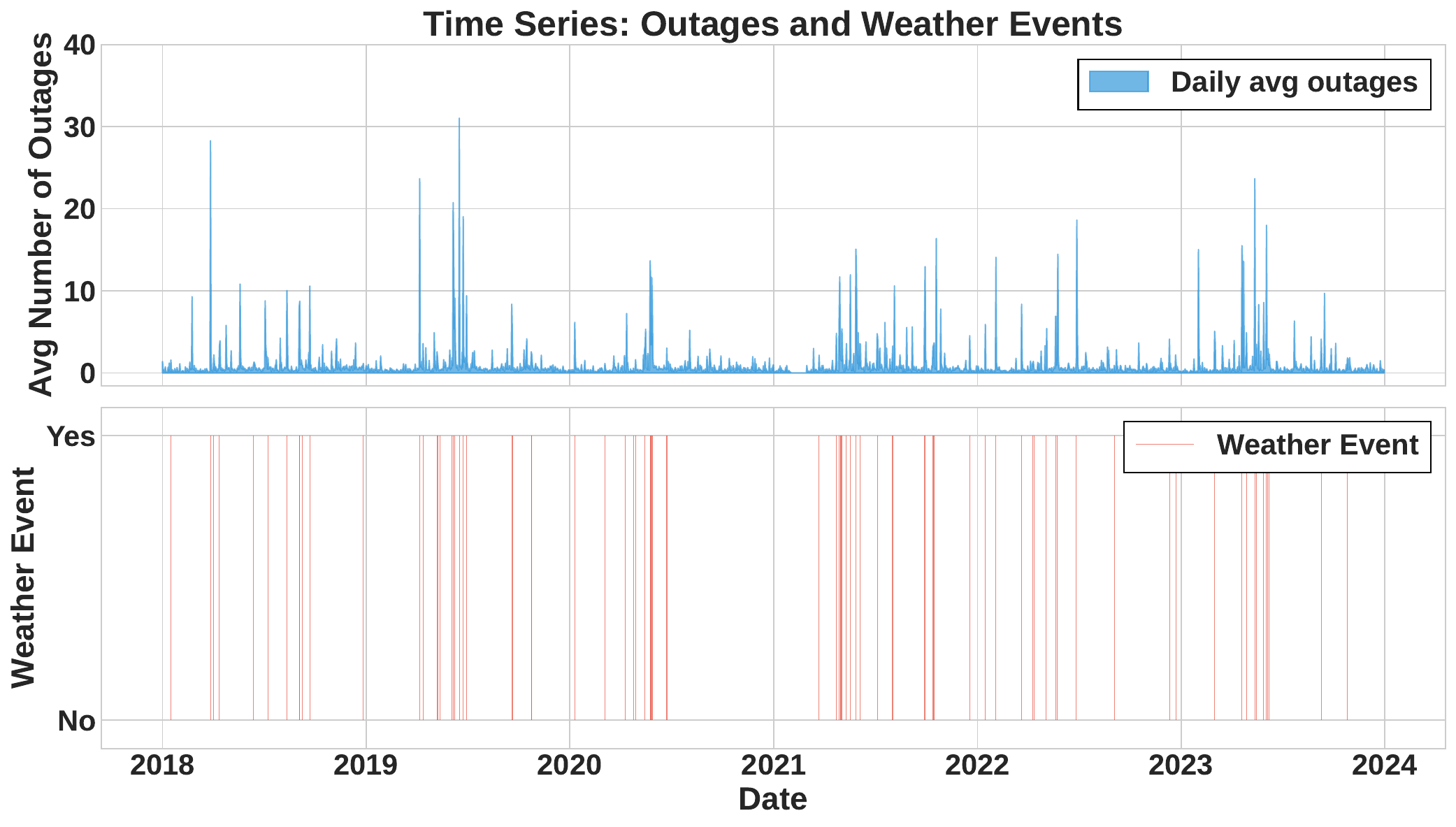}
  \caption{Daily average outages (blue) with NOAA-annotated severe weather days (red) for the central Texas service area in 2018-2023.}
  \label{fig:outage_timeseries}
\end{figure}

Before evaluating the use of LLMs for outage prediction using only weather data, we first verify that a clear and measurable relationship exists between the weather conditions and the outage records observed in the study area. This section establishes this relationship statistically before the predictive experiments are presented. To quantify the association between weather and power outages, we used two data sources. The first is a real-world dataset of historical outage records from a major Texas utility. For our analysis, we have excluded catastrophic events such as the Winter Storm Uri (Feb.\ 2021, Figure~\ref{fig:feb_2021_uri}). The second dataset is the National Oceanic and Atmospheric Administration (NOAA) Storm Events database \parencite{noaa_storm_events}, which provides official records of severe weather events in the area of interest, including event type, start time, and end time. The NOAA dataset was necessary because the numerical weather variables alone (temperature, wind speed, etc.) do not reliably indicate whether conditions are severe enough to cause faults. The storm event records provide a direct, verified label for when the weather was actually severe. Figure~\ref{fig:outage_timeseries} shows the daily average outage counts (blue) over the full study period with weather-event days (red) overlaid, illustrating both the variability of outages and their co-occurrence with recorded weather events.
Each NOAA event was expanded to hourly resolution and matched against the hourly outage counts. A $\pm $ 2-hour window was applied around each event, so that outages occurring slightly before or after the recorded weather event window were still captured because we can't rely on the precision of the NOAA dataset. Hours with no nearby weather events served as normal-weather records.

\begin{figure}[h]
  \centering
  \includegraphics[width=0.9\linewidth]{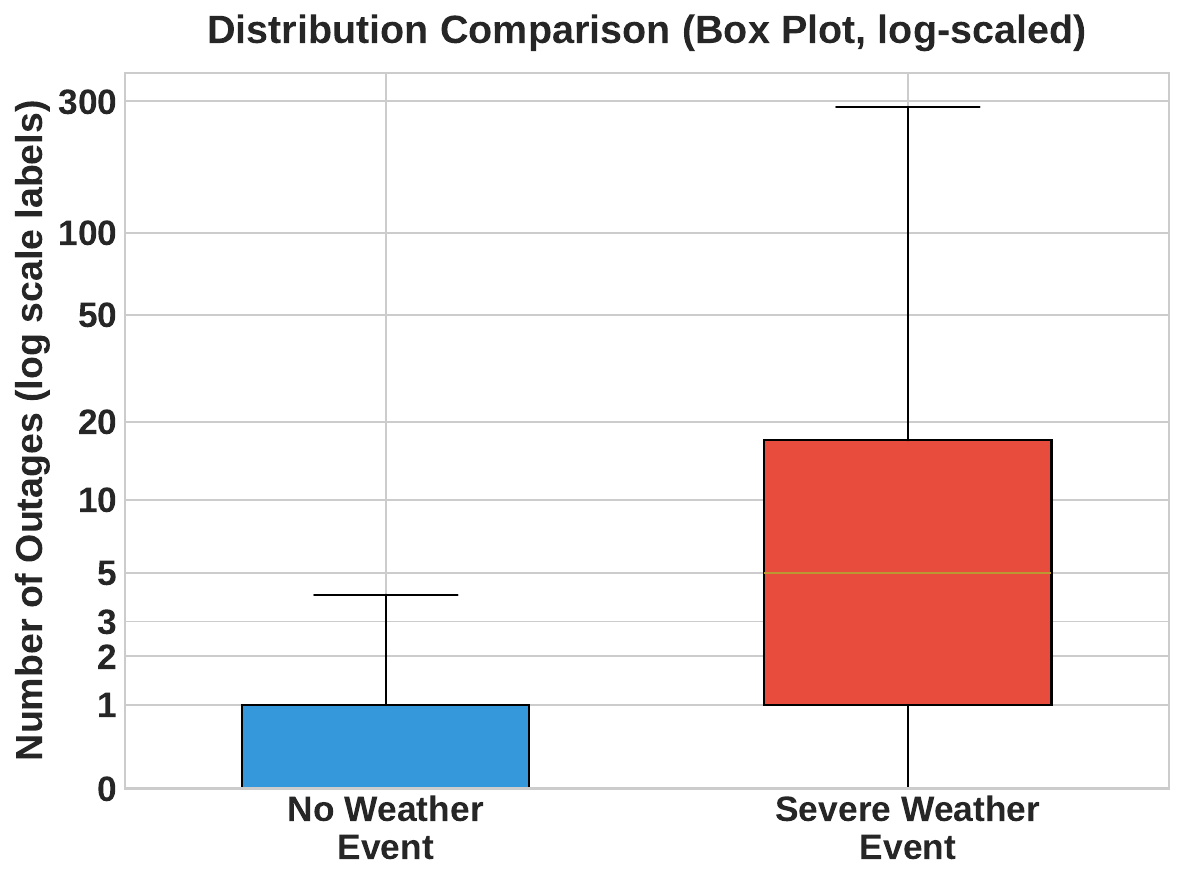}
  \caption{Hourly outage counts (log scale).}
  \label{fig:outage_boxplot}
\end{figure}

The association between weather events and forced outages is strong and statistically significant. On average, hours near a weather event had 2{,}082\% more outages than normal-weather hours (Mann--Whitney $U$ test, $p < 10^{-270}$). Figure~\ref{fig:outage_boxplot} compares the distribution of hourly outage counts between weather-event and normal-weather hours on a log scale, showing a clear upward shift in central tendency and a heavier upper tail during weather events. In the  service territory of interest, thunderstorm wind, lightning, and tornadoes showed the largest effects, with increases exceeding 3{,}500\% relative to the normal-weather case. All event types were significant at $p < 0.001$.

To go beyond correlation and assess causality, we applied two methods. First, lead-lag cross-correlation showed that weather activity at time $t$ predicts outages at $t+1\,\mathrm{h}$ ($r = 0.34$), while the reverse --- outages predicting future weather --- is substantially weaker (max $r = 0.25$). This directional asymmetry is consistent with weather causing outages rather than the reverse. Second, Granger causality tests confirmed that past weather values significantly improve the prediction of future outages ($F = 2{,}853$ at lag $1\,\mathrm{h}$, $p < 0.001$), with the forward direction approximately 19 times stronger than the reverse. Together, these results provide strong evidence that severe weather events directly drive power outages in the study area.

\section{Framing Outage Risk Prediction as a Severity Classification Problem}
\label{sec:FramingOutagePredictionSeverity}
Outage risk prediction helps utilities in several ways. It allows them to: a) warn customers early and provide reliable restoration estimates, b) position maintenance crews in advance rather than react after outages occur, and c) support informed decision-making for operational actions and mitigation measures \parencite{baembitov2024sensitivity}. There are several ways to define outage risk. Many works define it as the probability of having at least one outage at a given spatio-temporal data point \parencite{daniel_hicss, rashid_interpretability}. However, these approaches do not capture severity, since the implications differ markedly when, for example, only a few outages occur in a service area compared to multiple outages in the same area over the same time interval.

\begin{figure}[t]
    \centering
    \begin{subfigure}[t]{\linewidth}
        \centering
        \includegraphics[width=\linewidth]{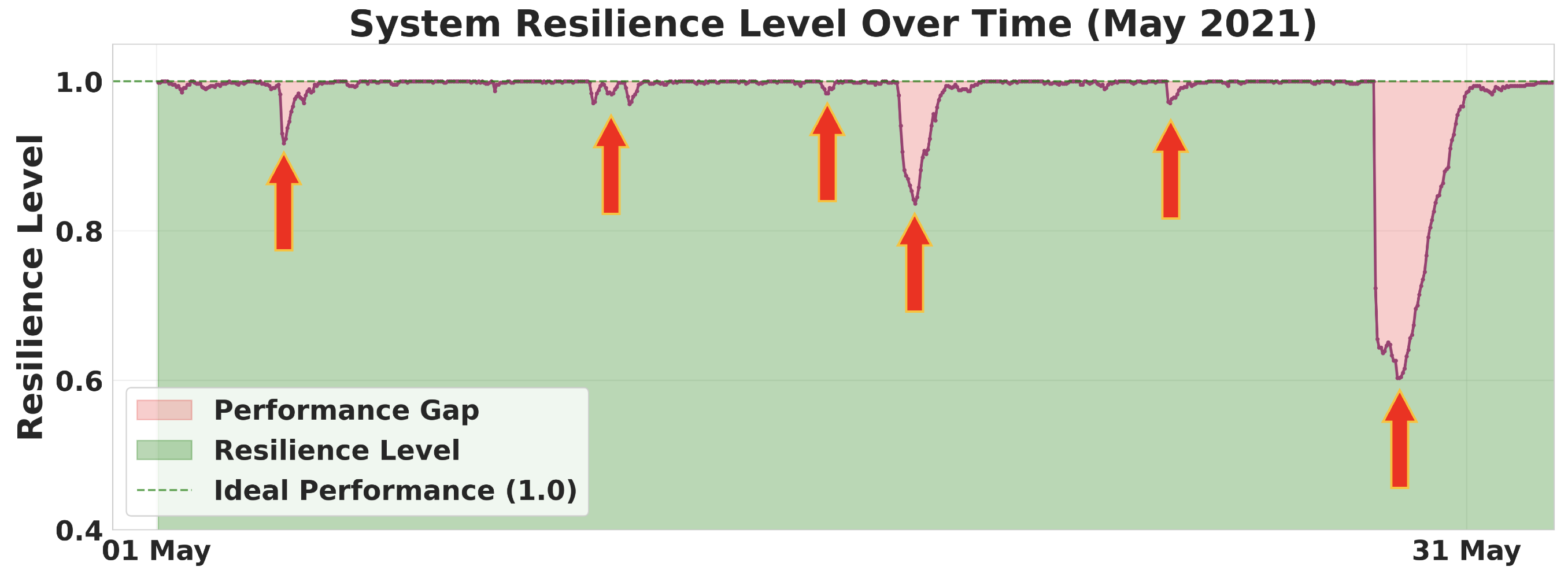}
        \caption{Red arrows indicate periods with recorded outage events.}
        \label{fig:may_2021_resilience}
    \end{subfigure}

    \vspace{1em}

    \begin{subfigure}[t]{\linewidth}
        \centering
        \includegraphics[width=\linewidth]{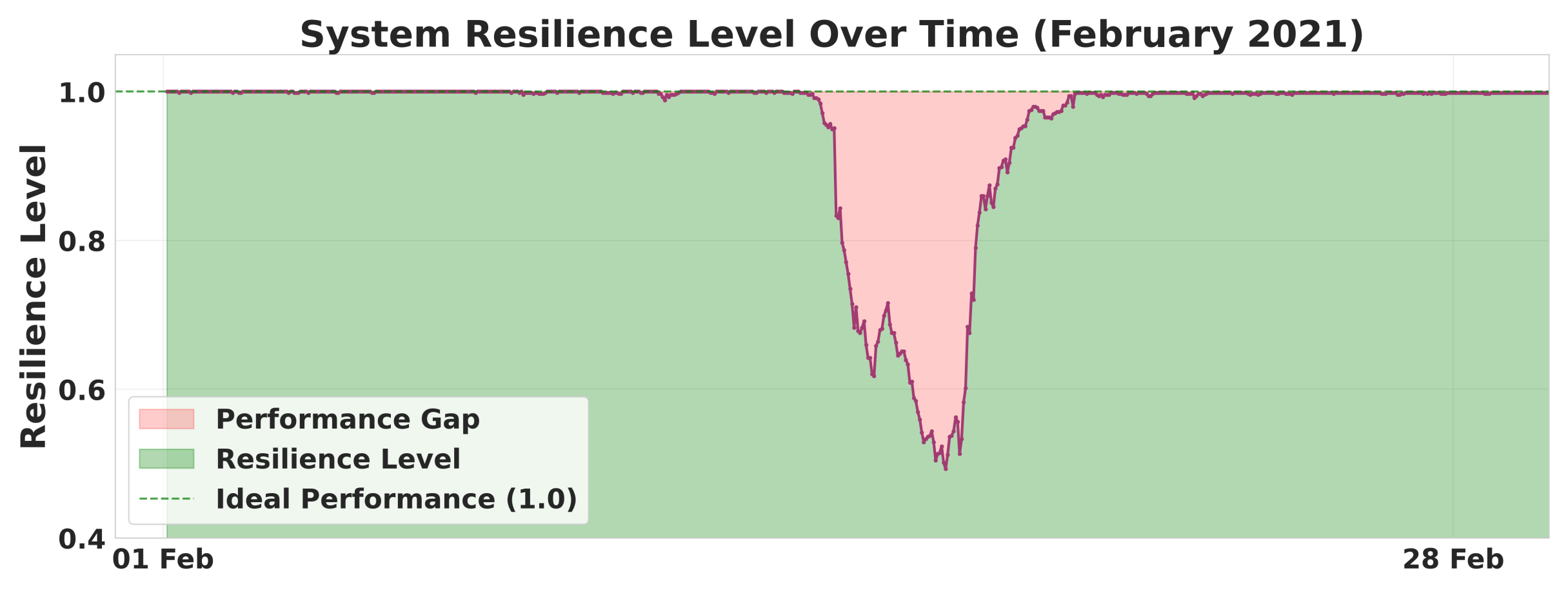}
        \caption{Winter Storm Uri.}
        \label{fig:feb_2021_uri}
    \end{subfigure}
    \caption{System resilience level in May (top) and February (bottom) 2021 for the service area in central Texas.}
    \label{fig:resilience_comparison}
\end{figure}

Figure~\ref{fig:may_2021_resilience} illustrates this point for the service territory of a utility in central Texas. Each red  arrow marks a period during which outages were recorded, yet the  corresponding drops in system resilience differ substantially. Some events cause only minor deterioration in resilience, while others, such as the one near the end of May, drive resilience level\footnote{The resilience level formula used in these plots is defined in the Appendix.} below 0.6 for an extended period. A risk definition based solely on outage occurrence would treat all these events as equivalent, not reflecting the operational reality that some events require far greater response than others. The importance of capturing severity becomes even clearer at the scale of major weather events. Figure~\ref{fig:feb_2021_uri} shows the system resilience level for the same service territory during February 2021, when Winter Storm Uri struck large portions of the Central United States. The event caused a prolonged and severe collapse in resilience, leaving millions of customers without power for days and resulting in widespread economic and human losses. A risk signal that only indicates whether an outage occurred cannot distinguish such an event from routine outage activity, even though the impact on operations and society is far greater. This motivates a definition of outage risk that also captures severity.

\section{Data Modeling}
\label{sec:data_model}
\textbf{Outage Data.} The raw outage dataset consists of individual outage records, one row per event, with fields for the reporting and the restoration timestamp. Each row represents a single outage event on a single feeder. To construct a structured time series suitable for machine learning, we aggregate these individual records into fixed-length, non-overlapping windows of $h$ hours. For each window boundary $t$, we count the number of outage events whose reporting timestamp falls within the interval $[t,\, t + h)$. This produces a regularly aligned hourly series, where each row corresponds to a window, and the target value is the total number of outages that will occur in the next $h$ hours, as observed from the current timestep $t$. For example, within a 3-hour window, the target value at 06:00 is the total number of outages reported between 06:00 and 09:00 --- the quantity the model is asked to predict at the time of the forecast.

\textbf{Target Variable Construction.} The outage count is a continuous, highly skewed variable — the vast majority of hours record zero or very few outages, while a small fraction of hours exhibit large spikes. To make the prediction task tractable and operationally meaningful, we convert the raw count into a binary \emph{high / low} risk label: an hour is classified as high risk (1) if the total number
of outages in the upcoming prediction window exceeds a threshold, and as low risk (0) otherwise. The threshold is defined with respect to the prediction window length, not as a single global value. For each window length --- 3, 6, or 12 hours ahead --- the threshold was chosen to yield a positive-class rate of approximately 5\% (95th quantile) across the dataset, which corresponds roughly to the upper tail of the outage count distribution for that window length. The resulting thresholds are: $\tau_3 = 5$ outages for the 3-hour window, $\tau_6 = 8$ outages for the 6-hour window, and $\tau_{12} = 10$ outages for the 12-hour window.

This window-dependent design reflects the physical reality of the problem. A 3-hour prediction window captures short events, so a moderate outage count (e.g., 5 outages) already represents an elevated stress condition within that short period. A 12-hour window, by contrast, spans multiple weather patterns and accumulates outages over a longer horizon, so a higher absolute threshold (ten outages) is required to identify a relatively severe event. In other words, the same raw count of five outages carries different significance depending on the window: it is a meaningful spike over 3 hours, but it is not enough to be classified as \emph{high} risk over 12 hours. Therefore, the high-risk label consistently captures the hours with the highest outage activity across all forecast horizons, regardless of how many outages a longer window naturally accumulates.

\textbf{Weather data.} Following prior work \parencite{daniel_hicss}, the weather data used in this study was downloaded from Open-Meteo, an open-source weather API that provides historical reanalysis data at hourly resolution. For each hour from 2018 to 2023, weather observations were retrieved across a 10$\times$10 spatial grid covering the service area. Because individual grid points may not be representative of the full area, each hourly timestep is summarized with spatial statistics across all 100 grid cells: the mean, minimum, maximum, and standard deviation. This produces a set of spatially aggregated features per hour, covering temperature, relative humidity, dew point, precipitation, wind speed, and wind gust at multiple heights, atmospheric pressure, cloud cover at multiple heights, soil moisture at multiple depths, vapor pressure deficit, evapotranspiration, and shortwave radiation. This spatial aggregation approach is consistent with prior work in the field \parencite{aljurbua2025spatiotemporal, rashid_interpretability, petridiscost}.

Since the prediction target spans a window of $h \in \{3, 6, 12\}$ hours, the weather features are constructed in one of two ways, depending on whether forecast weather information is available. When forecast weather is used, the $h$ hourly observations within $[t, t+h)$ are aggregated into a single feature vector using variable-specific rules: cumulative quantities such as precipitation and radiation are summed, extreme-sensitive variables such as wind gusts and pressure are taken as the window maximum or minimum, and continuous state variables such as temperature and humidity are averaged and statistics are calculated. The pressure range within the window  (e.g., the difference between the maximum and minimum) is also included as a feature, capturing rapid pressure drops associated with storm onset.
When forecast weather is not used, only the observed spatial weather statistics at the current timestep $t$ are provided as input, reflecting the conditions known at the moment of prediction.

\section{Predictive Modeling}
\label{sec:predictive_model}
To evaluate the hypothesis and assess the efficacy of the proposed methodology, we undertake a series of experiments encompassing various model configurations. These configurations include both machine learning models and large language models (LLMs). This experimental setup enables a comparative analysis of the predictive capabilities of traditional approaches versus LLM-based methods and evaluates the potential of the latter to attain competitive performance in zero-shot inference. Table~\ref{tab:class_distribution} provides a summary of the class distribution over six years for three forecast horizons. The datasets exhibit significant imbalance, by design though, with roughly 5\% of the instances labeled as high risk across all horizons. The rate of positive cases remains relatively consistent over the years, with 2021 showing the highest proportion across all horizons, even after excluding February (Winter Storm Uri).


\subsection{Predictive Modeling using LLMs}
\label{sec:PredictiveLLM}
We evaluate four LLMs in a zero-shot classification setting via the \text{OpenAI} API: GPT-4o, GPT-4o-mini, GPT-4.1, and GPT-4.1-mini. All four models receive an identical structured prompt comprising a system prompt that defines the task (estimating the probability that the upcoming $h$-hour window will experience a HIGH number of power outages) and a user prompt specifying the relevant weather conditions for that window. We evaluate two configurations: one in which the user prompt reports the weather forecast over the full prediction window, and another in which it reports only the current point-in-time observations. The models are instructed to respond with a JSON object containing two fields: \texttt{"probability"}, an integer between 0 and 100 representing the likelihood of a HIGH outage event, and \texttt{"reason"}, a short sentence explaining the key weather factors driving the assessment. Requiring the model to produce an explicit text to justify its prediction encourages intermediate reasoning over the input features, a mechanism shown to improve the accuracy and reliability of LLM outputs on complex inference tasks \parencite{wei2022chain, kojima2022large, zhang2024chain}. No examples or in-context demonstrations are provided to the LLMs; they rely entirely on their pre-trained knowledge to interpret the weather inputs and assess grid risk. We use a fixed threshold of 0.5 to convert each predicted probability into a binary decision, deliberately chosen rather than tuned, to keep the evaluation fully zero-shot, with no calibration based on the data.
\noindent
Below we present both the \textit{system} and \textit{user prompts} used for LLM inference in the weather-forecast configuration. The observed-weather configuration uses an identical structure, with the \textit{user prompt} referencing current conditions rather than the forecast window.
\begin{tcolorbox}[
  colback=gray!8,
  colframe=gray!40,
  boxrule=0.4pt,
  arc=2pt,
  left=6pt, right=6pt, top=4pt, bottom=4pt,
  fontupper=\ttfamily\small,
  breakable,
  coltitle=black,
  colbacktitle=gray!20,
]

\textbf{System prompt}\\[2pt]
You are an expert power grid reliability analyst for an electric utility service area.
Your task is to estimate the probability that the upcoming weather conditions will cause a HIGH number of power outages across the distribution grid of the service area in the next $h$ hours.\\[6pt]
A HIGH outage event means an unusually large spike in outages --- severe enough to represent a meaningful stress on the grid.
A LOW event means normal or near-normal conditions with few or no outages expected.\\[6pt]
Respond with ONLY a JSON object in this exact format:\\
\hspace*{1em}\texttt{\{"reason": "brief explanation ...", "probability": 85\}}\\[4pt]
The \texttt{"probability"} field must be an integer between 0 and 100 representing the likelihood of a HIGH outage event in the next $h$ hours.\\
The \texttt{"reason"} field must be a short sentence (max 30 words) explaining the main factors.\\
No extra text outside the JSON object.

\vspace{8pt}
\textbf{User prompt} (forecast mode)\\[2pt]
Weather conditions in the service area for the next $h$ hours:\\
... < weather information > ...\\
What is the probability of a HIGH outage event across the distribution grid of the service area in the next $h$ hours? Reply with your Reason and the Probability.

\end{tcolorbox}
\subsection{Predictive Modeling using ML}
We evaluated several supervised classifiers, including Random Forests, XGBoost, and a shallow Neural Network, but found that Logistic Regression and Gradient-Boosting decision trees (LightGBM) \parencite{nips_6449f44a} consistently achieved the best performance on this dataset and are therefore the models reported here. Logistic regression is trained with $\ell_2$ regularization ($C = 0.1$) and balanced class weights to account for the low positive-class rate. LightGBM is trained with both $\ell_1$ and $\ell_2$ regularization ($\alpha = 0.1$, $\lambda = 1.0$), and class imbalance is addressed by upweighting the minority class in proportion to the ratio of negative to positive training examples.

Following prior work \parencite{shen2025empowering}, machine learning models are evaluated using Leave-One-Year-Out (LOYO) cross-validation: in each fold, one calendar year is held out as the test set and the remaining years are used for training (or, in the case of the zero-shot LLM inference, simply skipped). This ensures that every model is evaluated on the same held-out years under identical conditions, enabling a fair comparison across paradigms. LOYO is well-suited to this setting because weather-driven outage patterns vary meaningfully from year to year due to differences in storm activity, seasonal anomalies, and grid conditions; holding out a full year, therefore, tests whether a model generalizes across these annual variations rather than overfitting to the specific patterns present in a particular subset of the data.

\begin{table}[h]
\centering
\caption{Class distribution per forecast horizon and year (high risk /
total timesteps). February 2021 (Winter Storm Uri) is excluded from all splits.}
\label{tab:class_distribution}
\setlength{\tabcolsep}{4pt}
\renewcommand{\arraystretch}{1.05}
\resizebox{\columnwidth}{!}{%
\begin{tabular}{lrrrrrr}
\toprule
 & \multicolumn{2}{c}{\textbf{3-hour}}
 & \multicolumn{2}{c}{\textbf{6-hour}}
 & \multicolumn{2}{c}{\textbf{12-hour}} \\
\cmidrule(lr){2-3}\cmidrule(lr){4-5}\cmidrule(lr){6-7}
\textbf{Year} & High & High\% & High & High\% & High & High\% \\
\midrule
2018 & 139 & 4.8\%  & 69 & 4.7\% & 33 & 4.5\% \\
2019 & 136 & 4.7\%  & 78 & 5.3\% & 41 & 5.6\% \\
2020 & 104 & 3.6\%  & 55 & 3.8\% & 28 & 3.8\% \\
2021 & 171 & 6.3\%  & 94 & 7.0\% & 50 & 7.4\% \\
2022 & 102 & 3.5\%  & 55 & 3.8\% & 24 & 3.3\% \\
2023 & 145 & 5.0\%  & 75 & 5.1\% & 36 & 4.9\% \\
\midrule
\textbf{All} & \textbf{797} & \textbf{4.6\%} & \textbf{426} & \textbf{4.9\%} & \textbf{212} & \textbf{4.9\%} \\
\bottomrule
\end{tabular}%
}
\end{table}



\section{Experimental Results}
\label{sec:experimental_results}
We discuss the comparative results between LLM and ML along three axes, covering the effect of the forecast horizon, the role of weather information, and the performance.  Macro-F1, precision, and recall for all models across the three forecast horizons and two weather configurations are reported in Table~\ref{tab:results}.

\begin{table*}[t]
\centering
\caption{Model performance across forecast horizons and weather configurations.
Best value overall per column is shown in \textbf{bold}.}
\label{tab:results}
\small
\setlength{\tabcolsep}{3pt}
\begin{tabular}{l *{3}{c} | *{3}{c} | *{3}{c}}
\toprule
& \multicolumn{3}{c}{\textbf{3h}} & \multicolumn{3}{c}{\textbf{6h}} & \multicolumn{3}{c}{\textbf{12h}} \\
\cmidrule(lr){2-4} \cmidrule(lr){5-7} \cmidrule(lr){8-10}
\textbf{Model} & Macro-F1 $\uparrow$ & Precision $\uparrow$ & Recall $\uparrow$ & Macro-F1 $\uparrow$ & Precision $\uparrow$ & Recall $\uparrow$ & Macro-F1 $\uparrow$ & Precision $\uparrow$ & Recall $\uparrow$ \\
\midrule
\multicolumn{10}{l}{\textit{\shortstack[l]{With weather\\forecast}}} \\
\midrule
GPT-4o              & 0.57 & 0.15 & 0.56 & 0.57 & 0.15 & 0.65 & 0.55 & 0.14 & 0.82 \\
GPT-4o-mini         & 0.37 & 0.07 & \textbf{0.78} & 0.33 & 0.07 & \textbf{0.88} & 0.29 & 0.07 & \textbf{0.93} \\
GPT-4.1             & 0.60 & 0.29 & 0.20 & 0.63 & 0.29 & 0.31 & 0.63 & 0.25 & 0.41 \\
GPT-4.1-mini        & 0.61 & 0.27 & 0.26 & 0.63 & 0.24 & 0.40 & 0.61 & 0.20 & 0.60 \\
LightGBM            & \textbf{0.74} & \textbf{0.46} & 0.56 & 0.70 & 0.35 & 0.61 & \textbf{0.67} & \textbf{0.29} & 0.65 \\
Logistic Regression & 0.73 & 0.43 & 0.58 & \textbf{0.71} & \textbf{0.40} & 0.53 & 0.66 & \textbf{0.29} & 0.55 \\
\midrule
\multicolumn{10}{l}{\textit{\shortstack[l]{Without weather\\forecast}}} \\
\midrule
GPT-4o              & 0.48 & 0.09 & 0.61 & 0.49 & 0.09 & 0.57 & 0.50 & 0.10 & 0.58 \\
GPT-4o-mini         & 0.15 & 0.05 & \textbf{0.95} & 0.16 & 0.05 & \textbf{0.96} & 0.13 & 0.05 & \textbf{0.97} \\
GPT-4.1             & 0.58 & 0.18 & 0.27 & 0.59 & 0.20 & 0.23 & 0.57 & 0.17 & 0.22 \\
GPT-4.1-mini        & 0.61 & 0.23 & 0.31 & 0.61 & 0.23 & 0.29 & 0.58 & 0.18 & 0.26 \\
LightGBM            & 0.71 & \textbf{0.49} & 0.39 & 0.69 & 0.34 & 0.54 & \textbf{0.61} & \textbf{0.21} & 0.48 \\
Logistic Regression & \textbf{0.73} & 0.43 & 0.60 & \textbf{0.70} & \textbf{0.35} & 0.56 & 0.60 & 0.20 & 0.46 \\
\bottomrule
\end{tabular}
\end{table*}


\textbf{Effect of forecast horizon.} Performance generally degrades as the horizon increases, which is expected given the greater uncertainty in predicting outage risk further ahead. ML models see a more pronounced drop, with LightGBM macro-F1 falling from 0.74 (3h) to 0.67 (12h) in the with-forecast setting, and from 0.71 to 0.61 without forecast. LLMs are more stable across horizons, with GPT-4.1 and GPT-4.1-mini maintaining macro-F1 around 0.60-0.63 regardless of window length, suggesting that their predictions rely less on the specific aggregation of weather features and more on general meteorological reasoning.

\textbf{Role of weather information.} Providing a weather forecast (\textit{in the forecast} configuration) consistently improves performance for ML models. As shown in Table~\ref{tab:gain_and_horizon}, LightGBM gains +0.03 (+4.2\%) at 3h and +0.06 (+9.8\%) at 12h, while Logistic Regression is flat at 3h but gains +0.06 (+10.0\%) at 12h, suggesting that forecast features become increasingly valuable as the horizon extends. For LLMs, the picture is more varied. GPT-4o-mini shows the largest absolute gains (+0.22, +146.7\% at 3h), but these numbers are misleading. Without a weather forecast, the model achieves a recall of at least 0.95 by classifying nearly every window as high risk (precision $\approx$0.05), making its without-forecast baseline trivially low and any improvement over it uninformative. GPT-4.1 gains +0.02--+0.06 (3.4\%--10.5\%) across horizons, while GPT-4.1-mini is nearly unchanged at short horizons (at 3h) and gains only +0.03 (+5.2\%) at 12h. GPT-4o shows moderate gains of +0.05--+0.09 (10.0\%--18.7\%). Overall, ML models exploit forecast features more consistently, while LLMs appear to derive most of their predictive signal from current conditions alone and benefit only marginally from the additional forecast window. Importantly, access to the weather forecast never degrades any model's performance. All gains reported in Table~\ref{tab:gain_and_horizon} are either positive or zero, indicating that providing forecast information is always safe and, in many cases, beneficial.

\textbf{Machine Learning and LLM performance.} ML models outperform all LLMs on macro-F1 and precision across all settings. In the with-forecast configuration, LightGBM achieves the highest macro-F1 at 3h (0.74) and 12h (0.67), while Logistic Regression leads at 6h (0.71). The best-performing LLM is GPT-4.1-mini at 3h (0.61) and 6h (0.63), and GPT-4.1 at 12h (0.63 with forecast). The gap between the best ML model and the best LLM ranges from 0.04 to 0.13 macro-F1 points, depending on the horizon and configuration.

\textbf{Recall tells a different story.} GPT-4o achieves a recall of 0.82 at 12h with forecast, surpassing LightGBM (0.65) and Logistic Regression (0.55). However, this high recall comes at the cost of very low precision (0.14), indicating that GPT-4o tends to over-predict high-risk events. GPT-4o-mini exhibits an even more extreme pattern, with recall ranging from 0.78 to 0.97 across settings (above 0.90 in all without-forecast configurations) but precision near 0.05--0.07, effectively flagging nearly every window as high risk. ML models occupy a middle ground, with LightGBM and Logistic Regression achieving recall in the range of 0.39--0.65 across horizons and configurations, considerably lower than GPT-4o-series models, but paired with precision of 0.20--0.49, which is a far more balanced operating point for operational deployment. It is worth noting that raw predictive metrics do not fully capture the value of LLMs in this context. Unlike ML models, LLMs accompany each prediction with a human-readable rationale for the risk assessment, which may support operator decision making even when predictive accuracy falls short. The two paradigms are therefore not strictly comparable on accuracy alone.

\begin{table*}[t]
\centering
\caption{Macro-F1 gain with weather forecast ($\Delta$W, absolute) across forecast horizons.}
\label{tab:gain_and_horizon}
\small
\setlength{\tabcolsep}{6pt}
\begin{tabular}{l rrr}
\toprule
& \multicolumn{3}{c}{\textbf{Forecast gain} $\Delta$W (with $-$ without)} \\
\cmidrule(lr){2-4}
\textbf{Model} & 3h & 6h & 12h \\
\midrule
GPT-4o              & +0.09 (+18.7\%) & +0.08 (+16.3\%) & +0.05 (+10.0\%) \\
GPT-4o-mini         & +0.22 (+146.7\%) & +0.17 (+106.2\%) & +0.16 (+123.1\%) \\
GPT-4.1             & +0.02 (+3.4\%)  & +0.04 (+6.8\%)  & +0.06 (+10.5\%) \\
GPT-4.1-mini        & +0.00 (+0.0\%)  & +0.02 (+3.3\%)  & +0.03 (+5.2\%)  \\
LightGBM            & +0.03 (+4.2\%)  & +0.01 (+1.4\%)  & +0.06 (+9.8\%)  \\
Logistic Regression & +0.00 (+0.0\%)  & +0.01 (+1.4\%)  & +0.06 (+10.0\%) \\
\bottomrule
\end{tabular}
\end{table*}

\section{Discussion}
\label{sec:discussion}
The following discussion is organized along two dimensions. We first examine model performance through standard predictive metrics and then assess the results against the three design criteria (transferability, scalability, and interpretability) introduced in Section~\ref{sec:motivation}.

\subsection{Performance Metrics}
\textbf{Performance-wise, ML models remain the superior choice.} Logistic Regression and LightGBM are trained directly on the target distribution and learn which weather features are most predictive for this service region — an advantage that is most pronounced at shorter horizons, where forecast features provide a clear, immediate signal.


\textbf{The significance of LLM generation surpasses that of model size.} When comparing GPT-4o and GPT-4.1, the latter consistently outperforms the former on macro-F1 and precision, despite comparable parameter scales. The patterns observed in GPT-4o-mini and GPT-4.1-mini are similar: GPT-4.1-mini markedly outperforms GPT-4o-mini across all evaluated metrics, except for recall. Furthermore, GPT-4.1-mini exceeds GPT-4o in macro-F1 scores across all horizons and configurations (e.g., 0.61 versus 0.57 at 3 hours with forecast). This suggests that architectural and training improvements between LLM generations have a large effect.

\textbf{The recall-precision trade-off differs across LLMs.} The over-prediction behavior of GPT-4o and GPT-4o-mini has a direct operational cost: frequent false alarms erode operator trust and trigger unnecessary crew mobilizations. GPT-4.1 and GPT-4.1-mini take the opposite stance, trading recall for precision, which is a more sustainable operating point for day-to-day deployment. This divergence is consistent with evidence that different LLMs map perceived likelihood to inconsistent numeric probability values \parencite{petridis2026unlikely}, so a uniform 0.5 threshold does not correspond to the same underlying confidence level across models. The choice between them therefore depends on whether the operational priority is to catch every high-risk event or to minimize unnecessary responses.


\textbf{Weather forecasts have a limited effect on LLMs' performance.} While ML models clearly benefit from window-aggregated forecast features, LLMs show only modest improvement when switching from point-in-time to full-window forecast inputs. This may reflect how LLMs process tabular weather data. Rather than integrating multiple feature values in a fine-grained way, they appear to rely on a few salient indicators (e.g., wind speed, precipitation) that are already present in the snapshot. This finding has a practical implication: in deployment settings where real-time forecast data is unavailable or costly, LLMs can still provide reasonable risk estimates based solely on current observations.

\subsection{Design Criteria}

\textbf{Transferability.} A key motivation for LLMs was their potential to operate without labeled data from the target service area. The results support this: as reported in Section~\ref{sec:experimental_results}, GPT-4.1 and GPT-4.1-mini achieve macro-F1 scores of 0.57--0.63 with no exposure to historical outage records from this utility, relying entirely on pre-trained meteorological knowledge. The same prompt template can be applied to any service territory without retraining, data collection, or domain-specific model development — a clear practical transferability advantage over supervised models, which must be rebuilt from scratch for each new territory.

\textbf{Scalability.} LLMs demonstrate stable performance across all three forecast horizons and both weather input configurations. Adapting to a different horizon or a new geographic region requires only a prompt adjustment, with no changes to the model or any training pipeline. Furthermore, incorporating additional input signals --- such as vegetation indices or historical outage context --- requires no retraining or feature engineering; it is purely a matter of prompt design. This stands in contrast to supervised models, which require a full retraining cycle whenever the input feature set changes.

\textbf{Interpretability.} Every LLM prediction in our experiments was accompanied by a natural-language explanation of the driving factors (presented in Section \ref{sec:motivation}), a capability that supervised models do not provide. Critically, this benefit holds even when LLM accuracy falls short of ML models — an explanation of moderate quality still supports operator decision-making in a way that a bare risk score cannot. The results therefore suggest that ML and LLM approaches should not be viewed as direct competitors but as tools with complementary strengths: ML models deliver superior predictive performance, while LLMs provide useful reasoning that supports operator trust and informed decision-making.

\section{Conclusion}
\label{sec:conclusion}
This paper presented an initial assessment of potential use of large language models (LLMs) for zero-shot, weather-related forced outage risk prediction at the distribution grid. The findings indicate that, although supervised machine learning (ML) models continue to possess an advantage in predictive performance, more recent LLMs attain comparable performance in the absence of labeled training data and provide additional benefits in transferability, scalability, and interpretability across different environments. Rather than viewing these as mutually exclusive options, the two paradigms should be regarded as complementary—ML delivering precise risk assessments, and LLMs offering actionable explanations to aid operator decision-making. Several directions remain open for future work. Combining weather observations and forecasts in a single prompt, and enriching it with location identifiers, calendar information, and historical outage context, may substantially improve LLM accuracy. Evaluating open-source models such as LLaMA or Mistral \parencite{livebench} would reduce API costs and allow utility operators to share internal data without privacy concerns. Finally, an agentic AI formulation in which a model actively retrieves relevant information at inference time represents a particularly promising direction. To the best of our knowledge, this is the first study of its kind, and we hope it serves as a starting point for further research in this emerging area.

\printbibliography

\appendix
\section*{Appendix}
\label{app:resilience}
\noindent
\textbf{System Resilience Level.} The system resilience level at each point in time is defined as the fraction of feeders remaining operational:

\begin{equation}
    \phi(t) = \frac{F_{\text{op}}(t)}{F_{\text{total}}}
\end{equation}

\noindent where $F_{\text{op}}(t)$ is the number of operational feeders at time $t$ and $F_{\text{total}}$ is the total number of feeders in the system. The aggregate Resilience Index $R$ over an event window $[t_e,\, t_f]$ is computed via a trapezoid loss area:

\begin{equation}
    A = \sum_{i} \bigl(1 - \phi_i\bigr) \cdot \Delta t_i
\end{equation}

\begin{equation}
    R = 1 - \frac{A}{t_f - t_e}
\end{equation}

\noindent where $\phi_i$ is the resilience level during interval $i$, $\Delta t_i$ is the duration of that interval in hours, $t_e$ is the start of the first outage, and $t_f$ is the end of the last restoration. $R = 1$ indicates no degradation, while $R = 0$ indicates all feeders were offline for the entire event window.\\








\end{document}